\documentclass[sigconf, screen]{acmart}
\usepackage{extarrows}
\usepackage{color} 
\usepackage{xcolor}
\usepackage{makecell}
\usepackage{amsmath,amsfonts}
\usepackage{algorithmic}
\usepackage{array}
\usepackage{graphicx}
\usepackage{stfloats}
\usepackage{multicol}
\usepackage{multirow}
\usepackage{hhline}
\usepackage{colortbl}
\usepackage{bm}
\usepackage{url}
\usepackage{enumitem}
\usepackage{arydshln}
\usepackage{bbding}
\usepackage{utfsym}
\usepackage{makecell}
\usepackage{booktabs} 

\usepackage[most]{tcolorbox}
\definecolor{lightgray}{rgb}{0.88, 0.92, 0.98}
\definecolor{defblue}{rgb}{0.1843, 0.3333, 0.6}
\definecolor{defred}{rgb}{0.88, 0.2510, 0.3294}

\definecolor{defgreen1}{rgb}{ 0.910,  0.953,  0.855}
\definecolor{defgreen2}{rgb}{0.82,  0.902,  0.710}
\definecolor{defgreen3}{rgb}{0.713,  0.903,  0.648}
\definecolor{defgreen4}{rgb}{ 0.725,  0.855,  0.561}
\definecolor{defred1}{rgb}{ 0.98,  0.855,  0.871}
\definecolor{defred2}{rgb}{ 1,  0.798,  0.799}
\definecolor{defred3}{rgb}{ 1,  0.398,  0.399}

\definecolor{defyellow}{rgb}{1,  0.983,  0.717}
\definecolor{defyellowtext}{rgb}{1,  0.851,  0.438}

\AtBeginDocument{%
  \providecommand\BibTeX{{%
    \normalfont B\kern-0.5em{\scshape i\kern-0.25em b}\kern-0.8em\TeX}}}

\copyrightyear{2025} 
\acmYear{2025} 
\setcopyright{acmlicensed}\acmConference[MM'25]{Make sure to enter the correct
  conference title from your rights confirmation email}{October 27 - October 31,
  2025}{Dublin, Ireland.}
\acmBooktitle{Proceedings of the 33rd ACM International Conference on Multimedia (MM'25), October 27- October 31, 2025, Dublin, Ireland}
\acmDOI{xxxxxxxxxxxxxxx}
\acmISBN{xxx-x-xxxx-xxxx-x/xx/xx}

\acmSubmissionID{xxx}

\setcopyright{none}
\renewcommand\footnotetextcopyrightpermission[1]{}
\begin{document}

\title{OFFSET: Segmentation-based Focus Shift Revision for \\ Composed Image Retrieval}

\author{Zhiwei Chen}
\affiliation{
  \institution{\normalsize Shandong University}
  \city{Jinan}
  \country{China}
  }
\email{zivczw@gmail.com}

\author{Yupeng Hu}
\authornote{Corresponding author: Yupeng Hu.}
\affiliation{%
  \institution{\normalsize Shandong University}
  \city{Jinan}
  \country{China}
  }
\email{huyupeng@sdu.edu.cn}

\author{Zixu Li}
\affiliation{
  \institution{\normalsize Shandong University}
  \city{Jinan}
  \country{China}
  }
\email{lizixu.cs@gmail.com}

\author{Zhiheng Fu}
\affiliation{
  \institution{\normalsize Shandong University}
  \city{Jinan}
  \country{China}
  }
\email{fuzhiheng8@gmail.com}

\author{Xuemeng Song}
\affiliation{
  \institution{\normalsize City University of Hong Kong}
  \city{Hong Kong}
  \country{China}
  }
\email{sxmustc@gmail.com}

\author{Liqiang Nie}
\affiliation{%
  \institution{\normalsize Harbin Institute of Technology (Shenzhen)}
  \city{Shenzhen}
  \country{China}
  }
\email{nieliqiang@gmail.com}

\begin{abstract}
Composed Image Retrieval (CIR) represents a novel retrieval paradigm that is capable of expressing users' intricate retrieval requirements flexibly. It enables the user to give a multimodal query, comprising a reference image and a modification text, and subsequently retrieve the target image. Notwithstanding the considerable advances made by prevailing methodologies, CIR remains in its nascent stages due to two limitations: 1) inhomogeneity between dominant and noisy portions in visual data is ignored, leading to query feature degradation, and 2) the priority of textual data in the image modification process is overlooked, which leads to a visual focus bias. To address these two limitations, this work presents a focus mapping-based feature extractor, which consists of two modules: dominant portion segmentation and dual focus mapping. It is designed to identify significant dominant portions in images and guide the extraction of visual and textual data features, thereby reducing the impact of noise interference. Subsequently, we propose a textually guided focus revision module, which can utilize the modification requirements implied in the text to perform adaptive focus revision on the reference image, thereby enhancing the perception of the modification focus on the composed features. The aforementioned modules collectively constitute the segmentatiOn-based Focus shiFt reviSion nETwork (\mbox{OFFSET}), and comprehensive experiments on four benchmark datasets substantiate the superiority of our proposed method. 
The codes and data are available on~\href{https://zivchen-ty.github.io/OFFSET.github.io/}{https://zivchen-ty.github.io/OFFSET.github.io/}.
\end{abstract}



\maketitle

\section{Introduction}\label{sec:intro}

\begin{figure}[t]
\begin{center}
\includegraphics[width=0.95\linewidth]{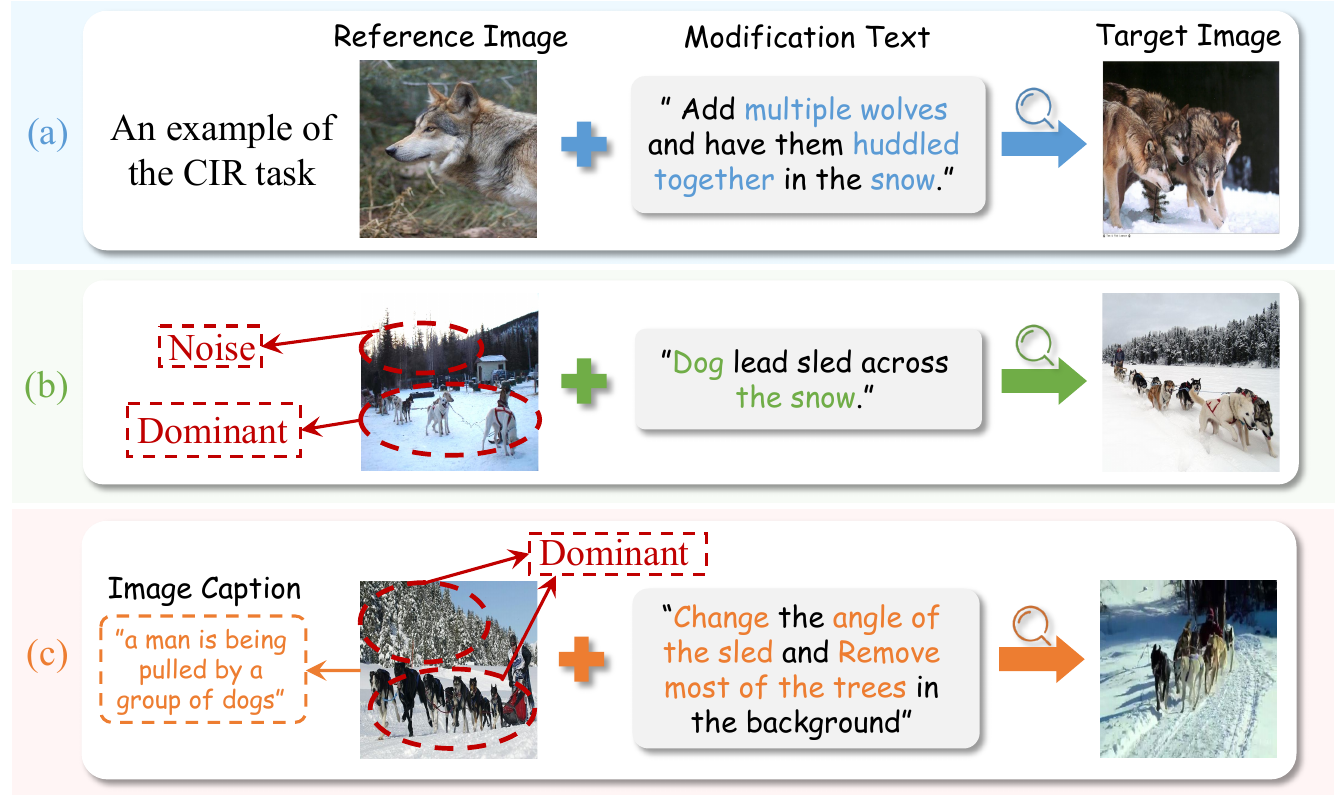}
\end{center}
\vspace{-12pt}
   \caption{(a) gives an example of the CIR task. (b) demonstrates the phenomenon of inhomogeneity in visual samples, where images frequently comprise dominant and noisy regions. (c) illustrates the advantages of applying text-priority during multimodal feature composition. The image caption treats ``trees'' as background noise information, which is inconsistent with the focus on modification text and may result in inaccurate composition results. However, when modification text is the primary objective, ``trees'' can be re-identified as the dominant region, thereby facilitating the construction of more accurate composed features.}
   \vspace{-12pt}
\label{fig:intro}
\end{figure}

In recent years, with the development of multimodal learning techniques~\cite{wu2025evaluation,bi2024visual,liu2025setransformer,bi2025prism,wu2025conditional,huang2023robust,liu2025gated, bi2025cot,tian2025core,wang2024explicit,tang2022youcan,huang2025enhancing}, there has been a growing interest in Composed Image Retrieval (CIR)~\cite{tirg} as a novel image retrieval paradigm~\cite{han2023fashionsap, encoder}. As illustrated in Figure~\ref{fig:intro}(a), the input of CIR is a multimodal query, comprising a reference image and a modification text. The modification text is used to convey the user's modification request for the reference image, and the model retrieves the target image based on the retrieval intent conveyed by the multimodal query. CIR's ability to express complex retrieval intent makes it a promising candidate for applications in areas such as information detection~\cite{qiu2025tab,zhang2024simultaneously,xu2025robustanomalydetectionnetwork,wang2025evaluating,xu2025hdnet,tang2022few}, image editing~\cite{lu2023tf,li2025set,lu2024robust,gao2024eraseanything,lu2024mace}, information prediction~\cite{wu2024novel,qiu2025duet,yu2025prnet,huang2024rock,qiu2025easytime,chen2024tokenunify} and object manipulation~\cite{zhou2024ssfold,huangexploring,zhou2025dual,huangnodes,zhou2025learningefficientroboticgarment}.

Nevertheless, despite the considerable advances made in previous research~\cite{dqu, encoder, song2024syncmask}, the field of CIR remains in its nascent stages, ignoring the following two phenomena.
1) \textbf{The visual inhomogeneity}. In the open domain, images often exhibit a complex visual composition, comprising both dominant regions that are strongly correlated with the modification requirements (\textit{e.g.}, ``dog'' and ``sled'' in the reference image of Figure~\ref{fig:intro}(b)) and irrelevant noise regions (\textit{e.g.}, ``tree'' in the reference image of Figure~\ref{fig:intro}(b)). However, previous work has failed to take into account the inhomogeneity of dominant and noisy regions. Treating them equally can lead to noise interference during the feature extraction, which in turn may cause a degradation of the query features and inaccurate feature composition results. In the fashion domain, although the image composition is relatively simple, users tend to modify only some regions of the image, which gives rise to the problem of inhomogeneity. 
2) \textbf{The text-priority in multimodal queries}. As illustrated in Figure~\ref{fig:intro}(c), based on the reference image alone, there is a tendency for individuals to direct their attention to the visual elements ``dog'', ``sled'' and ``person'', while the elements ``snow'' and ``tree'' may not be considered. Nevertheless, when the modification text is taken into account, it becomes evident that the ``tree'' is also an important visual feature, which belongs to the dominant regions. Previous research equally treats the reference image and the modification text and ignores the semantic priority of the modification text during the composition, which may result in a modification focus bias.

To address the aforementioned limitations, our objective is to implement feature extraction based on dominant region perception and textually guided focus revision. This will serve to enhance the performance of CIR. However, this is non-trivial due to the following three challenges. 
1) \textbf{Noise indicative signal absence}. To mitigate the impact of noise regions on feature extraction, it is necessary to identify dominant and noise regions in the image. However, there is no explicit noise supervision signal. Consequently, the primary challenge is to distinguish dominant and noisy regions in the absence of explicit supervision. 
2) \textbf{Entity role uncertainty}. As the saying goes, ``his honey, my arsenic.'' Thus, the roles of the same entity in different samples may vary considerably. For instance, in Figure~\ref{fig:intro}(b), the ``tree'' is situated within the noise region, whereas in Figure~\ref{fig:intro}(c), the ``tree'' is located within the dominant region. This illustrates the difficulty of performing uniform role modeling for different samples. In light of this, the second challenge is to develop an adaptive dominant region mining approach for guiding the feature extraction process. 
3) \textbf{Cross-modal semantic conflict}. As the modification text expresses the modification requirements for the reference image, there is likely to be a semantic conflict between them. Therefore, the third challenge is to address the conflicting complex semantics and complete the multimodal feature composition.

In order to address those challenges, we propose segmentati\textbf{O}n-based \textbf{F}ocal shi\textbf{F}t revi\textbf{S}ion n\textbf{ET}work, OFFSET for short, which implements feature extraction based on focus mapping and guides focus revision based on modification text to obtain multimodal composed features. Specifically, we first design a feature extractor, which consists of two modules, dominant portion segmentation and dual focus mapping. The dominant portion segmentation module utilizes the visual language model BLIP-2~\cite{li2023blip} to generate the image caption as a role-supervised signal, thus dividing the dominant and noisy regions and obtaining the dominant segmentation. The dual focus mapping module focuses on the extraction of multimodal features, and in particular, we design this module into two branches: \textit{Visual Focus Mapping (VFM)} and \textit{Textual Focus Mapping (TFM)}, and under the guidance of dominant segmentation, the two modules accomplish adaptive focus mapping on visual and textual data, respectively. Finally, we propose a textually guided focus revision module, which utilizes the modification requirements embedded in the textual feature to perform adaptive focus revision on the reference image, and compose the multimodal features based on modification focus perception enhancement. Extensive experiments on four benchmark datasets demonstrate the superiority of our approach.

In summary, our contributions include:
\begin{itemize}[leftmargin=8pt]
	\item We mine the phenomena of non-homogeneity and text dominance present in CIR tasks and design a solution based on focus mapping and text-guided focus revision.
	\item We propose a new CIR model OFFSET, which implements focus mapping-based feature extraction and textually guided focus revision through several well-designed components.
	\item Extensive experiments are conducted on four benchmark datasets to validate the effectiveness and superiority of the proposed model OFFSET and its components. The results demonstrate that our model achieves state-of-the-art performance.
\end{itemize}

\begin{figure*}
\centering
\vspace{-10pt}
\includegraphics[width=0.9\linewidth]{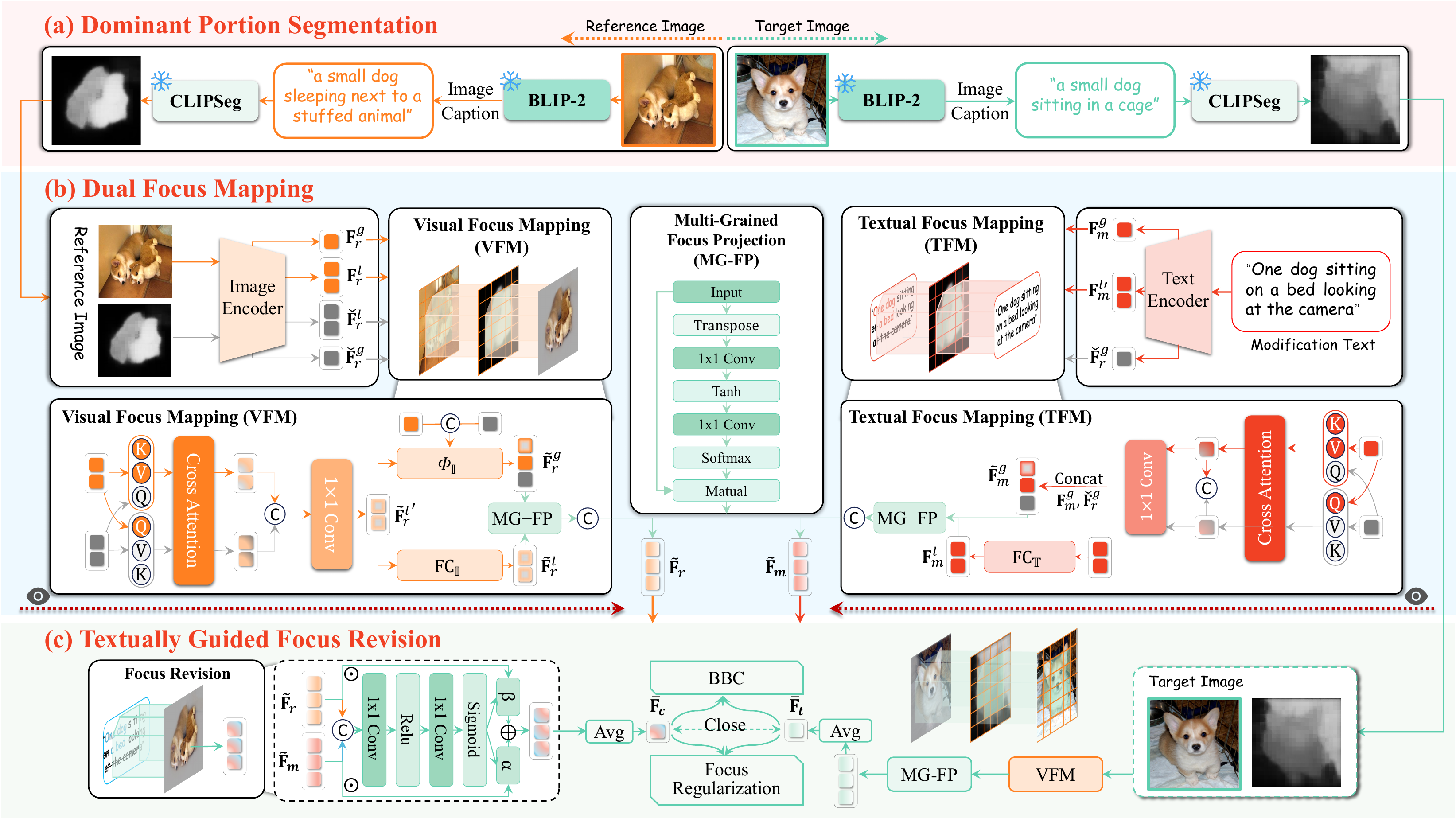}
\vspace{-10pt}
   \caption{The proposed OFFSET consists of three key modules: (a) Dominant Portion Segmentation, (b) Dual Focus Mapping, and (c) Textually Guided Focus Revision, where (a) and (b) collectively form the feature extractor.}
   \vspace{-13pt}
\label{fig:overview}
\end{figure*}

\section{Related Work}
Our work is closely related to Composed Image Retrieval (CIR) and semantic segmentation for mask generation.

\textit{\textbf{Composed Image Retrieval.}}
As a variant of image-text retrieval~\cite{chen2024bimcv}, this task aims to integrate the semantics of the reference image and the modification text to retrieve a target image, enabling flexible interaction and retrieval.  
According to the types of backbones used for feature extraction, current CIR approaches can be broadly classified into two categories, \textit{i.e.,} 1) conventional approaches based on traditional feature extraction models and 2) modern approaches utilizing VLP models for feature extraction. 
The first category typically extracts image and text features via traditional models, such as ResNet and LSTM, and then merges the corresponding multimodal query for retrieval~\cite{tirg,Chen_2020_CVPR,wen2021comprehensive}. 
Conversely, with the development of attention mechanism~\cite{wu2023towards,liu2025setransformer}, the second category~\cite{FineCIR,median,pair} utilizes vision and language pre-training (VLP) based models like CLIP~\cite{radford2021learning} for feature extraction. Notably, CLIP4Cir~\cite{baldrati2022conditioned} first fine-tunes CLIP and utilizes a simple \textit{Combiner} to achieve excellent performance, demonstrating a significant potential of VLP models. 
While these models achieve promising performance in CIR, they often overlook significant contextual focus clues during the composition, which shifts the visual focus and introduces irrelevant noise. To address it, our proposed model performs dual focus mapping, thereby enhancing the focus on the matching information between the image and text context.

\textit{\textbf{Semantic segmentation for mask generation.}}
Semantic segmentation~\cite{MSP-MVS,chen2023generative,zhao2022focal,yu2025crisp,SED-MVS,qian2024maskfactory,DVP-MVS,chen2023self,zhao2024guidednet,SD-MVS} has become a powerful technology for generating masks in various computer vision tasks. It is typically employed to create accurate object masks, which serve as crucial inputs for downstream tasks such as image editing, scene understanding, and object removal~\cite{tang2024textsquare,TSAR-MVS,tang2022optimal}. 
Conventionally, TransUNet~\cite{chen2021transunet} employs a hybrid architecture that combines a visual transformer encoder with a CNN-based decoder. 
In the NLP domain, DeepLabV3~\cite{bucher2019zero} synthesizes artificial, pixel-wise features for unseen classes based on word2vec label embeddings. 
Recently, VLP-based models have significantly impacted the field of semantic segmentation with the development of deep learning technology~\cite{tang2023character,huang2025final,zeng2025FSDrive,liu2025cpl,tang2024mtvqa,zeng2024driving}. Although the VLP models (\textit{e.g.,} CLIP~\cite{radford2021learning}) are not originally designed for segmentation, they have shown excellent zero-shot capabilities in generating object masks.
Building on this, methods such as CLIPSeg~\cite{luddecke2022image} further refine the process by prompt-guided segmentation, enabling more flexible and interactive mask generation based on the CLIP model.
This type of approach offers a promising avenue for mask generation in scenarios where solely limited training data is available or where there are unseen classes. Considering the flexible requirements of the CIR task, our model utilizes mask generation based on CLIPSeg to indirectly guide reference images and modification texts to focus on dominant portions.

\section{OFFSET}
As a primary novelty, our model (OFFSET) aims to perform dual focus mapping based on segmentation, followed by text-driven focus revision. As illustrated in Figure~\ref{fig:overview}, OFFSET consists of a focus mapping-based feature extractor and a textually guided focus revision module, where the feature extractor includes two modules: dominant portion segmentation and dual focus mapping. Specifically,
\mbox{(a) \textit{Dominant Portion Segmentation}} is applied to segment out the dominant portion of the image with the guidance of the image caption (detailed in Section~\ref{a.Dominant Portion Segmentation}).
(b) \textit{Dual Focus Mapping} utilizes the focus mapping in both visual and textual semantics based on the segmentation (described in Section~\ref{b.Dual Focus Mapping}). 
(c) \textit{Textually Guided Focus Revision} aims at revising the focus in the reference image to accurately identify the modification region based on the textual focus mapping results, and composing the modification requirements to match the target image (explained in Section~\ref{c.Textually Guided Focus Revision}). 
In this section, we first formulate the CIR task and then elaborate on each module.

\subsection{Problem Formulation}
OFFSET aims to address the challenging Composed Image Retrieval (CIR) task, whose goal is to retrieve the target image that satisfies the multimodal query. Let $\mathcal{T}=\{(x_{r},t_{m}, x_{t})_{n}\}_{n=1}^{N}$ denote a set of $N$ triplets, where $x_{r}, t_{m}$ and $x_{t}$ represent the reference image, modification text and target image, respectively. Inherently, we aim to learn a metric space where the embedding of the multimodal query ($x_{r}, t_{m}$) and corresponding target image $x_{t}$ ought to be as close as possible, which is formulated as, $\mathcal{H}\left(x_r, t_m\right) \rightarrow \mathcal{H}\left(x_t\right),$ where $\mathcal{H}$ denotes the to-be-learned embedding function for both multimodal queries and target images.

\subsection{Dominant Portion Segmentation}
\label{a.Dominant Portion Segmentation}
Initially, to exclude noise interference, we design the dominant portion segmentation for mining the image's primary area and generating the dominant segmentation, which lays the foundation for the subsequent \textbf{Focus Mapping}. Specifically, to accurately segment the image's primary area, we employ existing VLP models in image captioning, \textit{i.e.} BLIP-2~\cite{li2023blip}, to generate high-quality captions for both the reference image $x_{r}$ and the target image $x_{t}$. Formally, the caption process is formulated as,
\begin{equation}
\left\{
\begin{aligned}
&t_r=\operatorname{BLIP-2}\left(x_r\right),\\
&t_t=\operatorname{BLIP-2}\left(x_t\right),\\
\end{aligned}
\right.
    \label{blip-2}
\end{equation}
where $t_r$ and $t_t$ are the captions of the reference and target image, respectively. Subsequently, the captions and the corresponding images are input into CLIPSeg~\cite{luddecke2022image} to segment the dominant portion and the visual noise region. In formal terms, we have, 
\begin{equation}
\left\{
\begin{aligned}
&x^s_r=\operatorname{CLIPSeg}\left(x_r, t_r\right),\\
&x^s_t=\operatorname{CLIPSeg}\left(x_t, t_t\right),\\
\end{aligned}
\right.
    \label{ClipSeg}
\end{equation}
where $x^s_r$ and $x^s_t$ denote the dominant segmentation corresponding to the reference image $x_{r}$ and target image $x_{t}$, respectively, which highlights the image's dominant portion and simultaneously masks the noise information.

\subsection{Dual Focus Mapping}
\label{b.Dual Focus Mapping}
Afterwards, for mapping the focus in the dominant segmentation to both visual and textual data, we devise dual focus mapping, which is divided into two modules, Visual Focus Mapping (VFM) and Textual Focus Mapping (TFM). These two modules enable the visual and textual features to focus on the portions that are closely related to the multimodal queries. In the following, we subsequently present VFM and TFM.

\subsubsection{Visual Focus Mapping (VFM)} For VFM, since the reference image, target image, and dominant segmentation are modality-aligned images, we interact them at both the local and global levels for more accurate focus mapping. Specifically, taking the reference image ($x_r$) as an example (same as the target image), at the local level, we first extract local features by the image encoder of CLIP~\cite{radford2021learning}, which is also exploited in \textbf{TFM} to bridge visual and textual semantic information, formulated as follows,
\begin{equation}
\left\{
\begin{aligned}
&\textbf{F}^l_r=\varPhi_\mathbb{I}\left(x_r\right),\\
&\check{\textbf{F}}^l_r=\varPhi_\mathbb{I}\left(x_r^s\right),\\
\end{aligned}
\right.
    \label{Image Encode}
\end{equation}
where $\varPhi_\mathbb{I}$ denotes the penultimate layer of the image encoder, $ \textbf{F}^l_r, \check{\textbf{F}}^l_r\in \mathbb{R}^{C\times D_\mathbb{I}}$, $C$ is the visual channel number, and $D_\mathbb{I}$ is the visual embedding dimension. 
To map the focus information in the dominant segmentation to the reference image, we integrate their semantic information. Concretely, we utilize cross attention, where $\textbf{F}^l_r$ and $\check{\textbf{F}}^l_r$ are semantically interacted as each other's \textit{Query}, formulated as follows,
\begin{equation}
\left\{
\begin{aligned}
&\textbf{F}^{l(s)}_r=\operatorname{Cross\: Attention}\left(Q=\check{\textbf{F}}^l_r, \{K,V\}=\textbf{F}^l_r\right),\\
&\check{\textbf{F}}^{l(s)}_r=\operatorname{Cross\: Attention}\left(Q=\textbf{F}^l_r, \{K,V\}=\check{\textbf{F}}^l_r\right),\\
\end{aligned}
\label{cross_att}
\right.
\end{equation}
where $ \textbf{F}^{l(s)}_r, \check{\textbf{F}}^{l(s)}_r\in \mathbb{R}^{C\times D_\mathbb{I}}$. 
Then, with the receptive field of $\operatorname{1x1}$ convolution, we further integrate the specifics of the reference image and the dominant segmentation to obtain the focus-mapped local feature. And we align its dimension with CLIP's embedding dimension to facilitate the subsequent concatenation, which is formulated as follows,
\begin{equation}
\left\{
\begin{aligned}
&\tilde{\textbf{F}}^{l{\prime}}_r=\operatorname{1x1\: Conv}\left(\left[\textbf{F}^{l(s)}_r, \check{\textbf{F}}^{l(s)}_r\right]^\top\right)^\top,\\
&\tilde{\textbf{F}}^{l}_r=\operatorname{FC_{\mathbb{I}}}\left(\tilde{\textbf{F}}^{l{\prime}}_r\right),\\
\end{aligned}
\right.
\label{1x1_image}
\end{equation}
where $  \tilde{\textbf{F}}^{l{\prime}}_r\in \mathbb{R}^{C\times D_\mathbb{I}}$ and $  \tilde{\textbf{F}}^l_r\in \mathbb{R}^{C\times D}$ represent the focus-mapped local feature before and after dimension alignment, respectively, $D$ is CLIP's embedding dimension, and $\operatorname{FC_{\mathbb{I}}}$ is the fully connected layer.
Moreover, at the global level, to make full use of the fine-grained representation of dominant and noise portions, we obtain the focus mapping results at the global level with the local interaction information. Specifically, we feed the local features of the reference image, dominant segmentation and focus-mapped local features (\textit{i.e.} $\textbf{F}^l_r$, $\check{\textbf{F}}^{l}_r$ and $\tilde{\textbf{F}}^l_r$) into the last layer of CLIP image encoder $\varPhi_\mathbb{I}^g$ and concatenate their outputs, thus obtaining the focus-mapped global feature $\tilde{\textbf{F}}^{g}_r$, formulated as follows,
\begin{equation}
    \tilde{\textbf{F}}^{g}_r=\left[\textbf{F}^g_r,\: \check{\textbf{F}}^{g}_r, \:\tilde{\textbf{F}}^{g\prime}_r\right],
    \label{vis_global}
\end{equation}
where $ \textbf{F}^g_r=\varPhi_\mathbb{I}^g(\textbf{F}^l_r)$, 
$ \check{\textbf{F}}^{g}_r=\varPhi_\mathbb{I}^g(\check{\textbf{F}}^{l}_r)$, and $  \tilde{\textbf{F}}^{g\prime}_r=\varPhi_\mathbb{I}^g(\tilde{\textbf{F}}^{l}_r)$.
\mbox{Finally}, we align the global and local focus mapping results by \textbf{Multi-Grained Focus Projection (MG-FP)} (detailed in the section~\ref{MG-FP}) to obtain the reference focused feature $\tilde{\textbf{F}}_r$. Similarly, we can obtain the target focused feature $\tilde{\textbf{F}}_t$ \mbox{of the target image}.

\subsubsection{Textual Focus Mapping (TFM)} Since the modification text and the dominant segmentation belong to different modalities, the fine-grained semantic gap between them may cause focus mapping bias. Thus, we solely perform the textual focus mapping at the global level. Specifically, we first utilize the CLIP text encoder to extract the global feature $\textbf{F}^g_m$ of the modification text $t_m$. Meanwhile, to fully explore the multi-grained semantic information in the modification text, we also extract the local feature $\textbf{F}^{l{\prime}}_m$, formulated as follows,
\begin{equation}
\left\{
\begin{aligned}
&\textbf{F}^{l{\prime}}_m=\varPhi_\mathbb{T}\left(t_m\right),\\
&\textbf{F}^g_m=\varPhi_\mathbb{T}^g\left(\textbf{F}^{l{\prime}}_m\right),\\
\end{aligned}
\right.
    \label{Text Encode}
\end{equation}
where $ \varPhi_\mathbb{T}$ and $ \varPhi_\mathbb{T}^g$ denote the penultimate layer and last layer of the CLIP text encoder, respectively.
Then, similar to Eqn.$($\ref{cross_att}$)$, we utilize cross attention to semantically interact $ \textbf{F}^{g}_m$ and $\check{\textbf{F}}^{g}_r$ as each other's \textit{Query} to obtain the corresponding $\textbf{F}^{g(s)}_m, \check{\textbf{F}}^{g(s)}_r\in \mathbb{R}^{D}$. And then, we use $\operatorname{1x1}$ convolution to further integrate their specifics, \textit{i.e.} $\fontsize{8pt}{8pt}\tilde{\textbf{F}}^{g{\prime}}_m=\operatorname{1x1\: Conv}([\textbf{F}^{g(s)}_m, \check{\textbf{F}}^{g(s)}_r]^\top)^\top$, and concatenate it with the global feature of the modification text and the dominant segmentation (\textit{i.e.}, $\textbf{F}^{g}_m,\check{\textbf{F}}^{g}_r$). The focus-mapped global feature $\tilde{\textbf{F}}^{g}_m$ of the modification text is obtained as follows,
\begin{equation}
    \tilde{\textbf{F}}^{g}_m=\left[\textbf{F}^{g}_m,\check{\textbf{F}}^{g}_r,\tilde{\textbf{F}}^{g{\prime}}_m\right],
    \label{1x1_text}
\end{equation}
where $\tilde{\textbf{F}}^{g}_m\in \mathbb{R}^{3\times D}$. For the local feature of the modification text, similar to Eqn.$($\ref{1x1_image}$)$, we utilize the fully connected layer $\operatorname{FC_{\mathbb{T}}}$ for dimensional alignment, yielding $\textbf{F}^{l}_m\in \mathbb{ R}^{S\times D}$, where $S$ is the sequence channel number of modification text. Finally, we align the textual global and local focus mapping results via \textbf{MG-FP} (detailed in the section~\ref{MG-FP}) and obtain modification focused feature $\tilde{\textbf{F}}_m$.

\subsubsection{Multi-Grained Focus Projection (MG-FP)}
\label{MG-FP}
Since the above focus-mapping process employs multi-grained semantic information, the channel number of the mapping results varies across different granularities (\textit{e.g.}, the channel number of the focus-mapped global feature is 3, while that of the focus-mapped local feature is $C$ or $S$). It is not conducive to generating uniform focus mapping results due to the various focus areas of different channels. Thus, we design the MG-FP module to project the multi-grained and multi-channel focus mapping results into a unified focus channel. Specifically, taking the focus-mapped local feature $\tilde{\textbf{F}}^{l}_r$ of the reference image in \textbf{VFM} as an example, assuming that the number of uniform focus channels is $P$, we first resort the $\operatorname{1x1}$ convolution to obtain the projection weight $\textbf{W}_p$ of each focus channel, which is formulated as follows,
\begin{equation}
\textbf{W}_p=\operatorname{Softmax}\left(\operatorname{1x1\:Conv}\left(\operatorname{Tanh}\left(\operatorname{1x1\:Conv}\left({{}\tilde{\textbf{F}}^{l}_r}^{\top}\right)\right)\right)\right),
\label{projection_w}
\end{equation}
where $\textbf{W}_p\in\mathbb{R}^{P\times C}$. Subsequently, $\textbf{W}_p$ is aggregated \mbox{into $\tilde{\textbf{F}}^{l}_r$} to obtain the weighted focus-mapped local feature of the reference image $\tilde{\textbf{F}}^{l(f)}_r\in \mathbb{R}^{P\times D}$. Similarly, we can obtain the weighted focus-mapped global feature of the reference image $\tilde{\textbf{F}}^{g(f)}_r\in \mathbb{R}^{P\times D}$. Then, they are concatenated to obtain the reference focused feature $\tilde{\textbf{F}}_r\in \mathbb{R}^{2P\times D}$. Analogously, we can obtain target focused feature $\tilde{\textbf{F}}_t\in \mathbb{R}^{2P\times D}$ and \mbox{modification focused} feature $\tilde{\textbf{F}}_m\in \mathbb{R}^{2P\times D}$.

\subsection{Textually Guided Focus Revision} 
\label{c.Textually Guided Focus Revision}
As aforementioned, the lack of guidance on modification requirements results in inaccurate focus localization since the visual focus mapping results are modality-independent and solely contain the original visual information. Hence, we devise the textually guided focus revision module to utilize the textual modification semantic information for guiding the focus revision in the reference image and obtaining the multimodal composed feature, which is driven closer to the target focused feature. 

Specifically, we first perform a concatenated interaction between the reference focused feature and modification focused feature (\textit{i.e.}, $\tilde{\textbf{F}}_r, \tilde{\textbf{F}}_m$) via the $\operatorname{1x1}$ convolution, obtaining the revision weight $\textbf{W}_r$ for each focus channel, formulated as follows,
\begin{equation}
\textbf{W}_r\!\!=\!\operatorname{Sigmoid}\left(\operatorname{1x1\:Conv}\left(\operatorname{Relu}\left(\operatorname{1x1\:Conv}\left(\left[\tilde{\textbf{F}}_r, \tilde{\textbf{F}}_m\right]^{\top}\right)\right)\right)\right)^{\top}\!\!,
\label{revision_w}
\end{equation}
where $\textbf{W}_r\in \mathbb{R}^{P\times 2D}$. Afterwards, the chunk operation is utilized to split $\textbf{W}_r$ into $\alpha, \beta  \in \mathbb{R}^{P\times D}$, and aggregate them into $\tilde{\textbf{F}}_r, \tilde{\textbf{F}}_m$, respectively. Then, the aggregated features are summed to obtain the multimodal composed feature $\tilde{\textbf{F}}_c \in \mathbb{R}^{P\times D}$, formulated as follows,
\begin{equation}
\tilde{\textbf{F}}_c=\alpha\tilde{\textbf{F}}_r+\beta\tilde{\textbf{F}}_m.
    \label{combine}
\end{equation}

Subsequently, to push the multimodal composed feature $\tilde{\textbf{F}}_c$ closer to the target focused feature $\tilde{\textbf{F}}_t$, we employ batch-based classification loss~\cite{vo2019composing}, which is wildly utilized in CIR, formulated as follows,
\begin{equation}
\mathcal{L}_{rank} = \frac{1}{B} \sum_{i=1}^{B} -\log \left\{ \frac{\exp \left\{ \operatorname{s} \left( \bar{\textbf{F}}_{ci} , \bar{\textbf{F}}_{ti} \right)  / \tau\right\}}{ \sum_{j=1}^{B} \exp \left\{ \operatorname{s} \left( \bar{\textbf{F}}_{ci}, \bar{\textbf{F}}_{tj} \right) / \tau \right\}  } \right\},
\label{bbc}
\end{equation}
where $B$ is the batch size, $\bar{\textbf{F}}_{ci}, \bar{\textbf{F}}_{ti}\in \mathbb{R}^{D}$ represent the $i$-th average pooled $\tilde{\textbf{F}}_c, \tilde{\textbf{F}}_t$, respectively, $s(\cdot, \cdot)$ is the cosine similarity function, and $\tau$ is the temperature coefficient. 

Furthermore, we argue that $\tilde{\textbf{F}}_c, \tilde{\textbf{F}}_t$ require not only a high similarity degree but also a consistent distribution of focus degree in the focus channel, to fulfill the fine-grained focus matching between the multimodal composed feature and the target focused feature while improving the retrieval accuracy. Specifically, let $\mathbf{f}_i^{c}=\left[f_{i1}^{c},... ,f_{iB}^{c}\right]$ represents the focus degree distribution of the $i$-th multimodal composed feature in the batch, where $f_{ij}^{c}$ is the focus degree between $i$-th average pooled $\tilde{\textbf{F}}_c$ and $j$-th average pooled $\tilde{\textbf{F}}_t$, which can be computed as follows,
\begin{equation}
f_{ij}^{c} =  \frac{\exp \left\{ \operatorname{s} \left( \bar{\textbf{F}}_{ci} , \bar{\textbf{F}}_{tj} \right)  / \tau\right\}}{ \sum_{b=1}^{B} \exp \left\{ \operatorname{s} \left(\bar{\textbf{F}}_{ci}, \bar{\textbf{F}}_{tb} \right) / \tau \right\}  }.
\label{focus degree}
\end{equation}

Analogously, we obtain the focus degree distribution of the $i$-th target focused feature in the batch, denoted as $\mathbf{f}_i^{t}=[f_{i1}^{t},.... ,f_{iB}^{t}]$. Then, we define focus regularization to promote the convergence of the two focus degree distributions as follows,
\begin{equation}
\mathcal{L}_{fr}= \frac{1}{B}\sum_{i=1}^{B}D_{KL}\left({\mathbf{f}_i^{t}} \| {\mathbf{f}_i^{c}}\right)= \frac{1}{B}\sum_{i=1}^{B}\sum_{j=1}^{B} f_{ij}^{t} \log \frac{f_{ij}^{t}}{f_{ij}^{c}}.
\label{L_fr}
\end{equation}

Finally, we obtain the optimization function for OFFSET, 
\begin{equation}
    \mathbf{\Theta^{*}}=
    \underset{\mathbf{\Theta}}{\arg \min } \left( {\mathcal{L}}_{rank}+\mu {\mathcal{L}}_{fr} \right),
    \label{optimization}
\end{equation}
where $\mathbf{\Theta^{*}}$ is the to-be-learnt parameter for OFFSET and $\mu$ is the trade-off hyper-parameter.

\section{Experiments}
\subsection{Experimental Settings}

\subsubsection{Datasets}
Following previous works, we select three public datasets for evaluation, including two fashion-domain datasets, FashionIQ~\cite{wu2021fashion}, Shoes~\cite{guo2018dialog}, and an open-domain dataset CIRR~\cite{liu2021image}. 

\subsubsection{Implementation Details}
Following previous work~\cite{dqu}, OFFSET utilizes pre-trained CLIP (ViT-H/14 version) as the backbone. We trained \mbox{OFFSET} via AdamW optimizer with an initial learning rate of $1e$-$4$, while the learning rate for CLIP is set to $1e$-$6$ for better convergence and the batch size is $16$. We empirically set the embedding dimension $D$ = $1024$. Through \mbox{hyper-parameter} tuning, the focus channel number $P$ is set to $4$, and the temperature factor $\tau$ in Eqn.$($\ref{bbc},\ref{L_fr}$)$ is set to $0.1$ for all datasets. The trade-off hyper-parameter $\mu$ is searched via grid search and finally confirmed as $\mu\!=\!0.5$ on FashionIQ and Shoes, while $\mu\!=\!0.8$ for CIRR. All experiments were performed on a single NVIDIA A40 GPU with $48$GB memory and trained $10$ epochs.

\subsubsection{Evaluation.}
To ensure a fair assessment of model performance across different datasets, we adopt standard evaluation protocols following previous works~\cite{dqu,encoder}. The primary metric is Recall@$k$ (abbreviated as R@$k$). For the FashionIQ dataset, we utilize R@$10$, R@$50$, and their category-wise averages for the fair evaluation. For the Shoes dataset, we calculate R@$k$ ($k\!=\!1, 10, 50$) and their mean value. CIRR assessment included R@$k$ ($k\!=\!1, 5, 10, 50$), R$_{subset}$@$k$ ($k\!=\!1, 2, 3$), and the average of R@$5$ and R$_{subset}$@$1$.

\begin{table*}[h]
  \centering
      \caption{Performance comparison on FashionIQ relative to R@$k$($\%$). The overall best results are in bold, while the best results over baselines are underlined.}
      \tabcolsep=9pt
      \resizebox{0.9\linewidth}{!}{
    \begin{tabular}{l|cc|cc|cc|cc}
    \hline
    \hline
    \multicolumn{1}{c|}{\multirow{2}{*}{Method}}     & \multicolumn{2}{c|}{Dresses} & \multicolumn{2}{c|}{Shirts} & \multicolumn{2}{c|}{Tops\&Tees} & \multicolumn{2}{c}{Avg}  \\
\cline{2-9}         & R@$10$  & R@$50$  & R@$10$  & R@$50$  & \multicolumn{1}{c}{R@$10$} & R@$50$  & R@$10$  & R@$50$   \\ \hline 
\rowcolor[rgb]{ .949,  .949,  .949} \multicolumn{9}{c}{\textit{Traditional Model-Based Methods}}\\
    TIRG~\shortcite{tirg}~\textcolor{gray}{(CVPR'19)} & 14.87  & 34.66  & 18.26  & 37.89  & 19.08  & 39.62  & 17.40  & 37.39   \\
    VAL~\shortcite{Chen_2020_CVPR}~\textcolor{gray}{(CVPR'20)}& 21.12  & 42.19  & 21.03  & 43.44  & 25.64  & 49.49  & 22.60  & 45.04    \\
    CLVC-Net~\shortcite{wen2021comprehensive}~\textcolor{gray}{(SIGIR'21)}& 29.85  & 56.47  & 28.75  & 54.76  & 33.50  & 64.00  & 30.70  & 58.41   \\

    ARTEMIS~\shortcite{delmas2022artemis}~\textcolor{gray}{(ICLR'22)} & 27.16  & 52.40  & 21.78  & 43.64  & 29.20  & 54.83  & 26.05  & 50.29   \\
 
    MGUR~\shortcite{chen2022composed}~\textcolor{gray}{(ICLR'24)}& 32.61  & 61.34  & 33.23  & 62.55  & 41.40  & 72.51  & 35.75  & 65.47    \\
\hdashline
\rowcolor[rgb]{ .949,  .949,  .949} \multicolumn{9}{c}{\textit{VLP Model-Based Methods}}\\
    Prog. Lrn.~\shortcite{zhao2022progressive}~\textcolor{gray}{(SIGIR'22)}& 38.18  & 64.50  & 48.63  & 71.54  & 52.32  & 76.90 & 46.38  & 70.98   \\
    FashionSAP~\shortcite{han2023fashionsap}~\textcolor{gray}{(CVPR'23)} & 33.71  & 60.43  & 41.91  & 70.93  & 33.17  & 61.33  & 36.26  & 64.23  \\
    FAME-ViL~\shortcite{han2023fame}~\textcolor{gray}{(CVPR'23)} & 42.19  & 67.38  & 47.64  & 68.79  & 50.69  & 73.07  & 46.84  & 69.75  \\
    SyncMask~\shortcite{song2024syncmask}~\textcolor{gray}{(CVPR'24)}& 33.76  & 61.23  & 35.82  & 62.12  & 44.82  & 72.06  & 38.13  & 65.14  \\
    IUDC~\cite{iudc}~\textcolor{gray}{(TOIS'24)}& 35.22& 61.90 &41.86& 63.52& 42.19& 69.23& 39.76& 64.88 \\
    SADN~\cite{sadn}~\textcolor{gray}{(ACM MM'24)} &40.01 & 65.10 & 43.67 & 66.05 & 48.04 & 70.93 & 43.91 & 67.36 \\
    CaLa~\cite{cala}~\textcolor{gray}{(SIGIR'24)} &42.38&	66.08&	46.76&	68.16	&50.93&	73.42 &46.69 &	69.22 \\
    CoVR-2~\cite{covr-2}~\textcolor{gray}{(TPAMI'24)} &46.53& 69.60 &51.23 &70.64& 52.14 &73.27& 49.96 &71.17\\
    Candidate~\cite{candidate}~\textcolor{gray}{(TMLR'24)}& 48.14& 71.34& 50.15 &71.25& 55.23& 76.80& 51.17& 73.13\\
    SPRC~\cite{sprc}~\textcolor{gray}{(ICLR'24)}& 49.18 &	72.43 &	55.64 &	73.89 	&59.35 &	78.58 &	54.72 &	74.97 \\
    FashionERN~\cite{fashionern}~\textcolor{gray}{(AAAI'24)}  & 50.32  & 71.29  & 50.15  & 70.36  & 56.40 & 77.21  & 52.29  & 72.95 \\
    LIMN~\cite{limn}~\textcolor{gray}{(TPAMI'24)} & 50.72  & 74.52  & 56.08  & 77.09  & 60.94  & 81.85  & 55.91 & 77.82 \\
    LIMN+~\cite{limn}~\textcolor{gray}{(TPAMI'24)} &52.11 &75.21& 57.51& 77.92& 62.67& 82.66&57.43 &	78.60  \\
    DQU-CIR~\cite{dqu}~\textcolor{gray}{(SIGIR'24)}& \underline{57.63} &	\underline{78.56} 	&\underline{62.14} &	\underline{80.38} &	\underline{66.15} &	\underline{85.73} &	\underline{61.97} &	\underline{81.56} \\
    ENCODER~\cite{encoder}~\textcolor{gray}{(AAAI'25)}& 51.51& 76.95& 54.86& 74.93& 62.01& 80.88& 56.13& 77.59  \\
\rowcolor[rgb]{ .851,  .851,  .851}
    \multicolumn{1}{l|}{\textbf{OFFSET(Ours)}} & \textbf{\large 57.86} &	\textbf{\large 79.13} &	\textbf{\large 62.81} &	\textbf{\large 81.55} &	\textbf{\large 67.11} &	\textbf{\large85.87} & \textbf{\large62.59} &	\textbf{\large82.18}    \\
    \hline    \hline
    \end{tabular}
}
  \label{tab:fiq_shoes}%
\end{table*}%

\begin{table}[ht]
\centering
    \caption{Performance comparison on Shoes with respect to R@$k$($\%$). The overall best results are in bold, while the best results over baselines are underlined.}
    \resizebox{\linewidth}{!}{
       \begin{tabular}{l|ccc|c}
\hline\hline
    \multicolumn{1}{c|}{Method} & R@$1$   & R@$10$  & R@$50$  & Avg \\
    \hline
    \rowcolor[rgb]{ .949,  .949,  .949} \multicolumn{5}{c}{\textit{Traditional Model-Based Methods}}\\
    TIRG~\shortcite{vo2019composing}~\textcolor{gray}{(CVPR'19)}& 12.60  & 45.45  & 69.39  & 42.48  \\
    VAL~\shortcite{Chen_2020_CVPR}~\textcolor{gray}{(CVPR'20)}  & 17.18  & 51.52  & 75.83  & 48.18  \\
    CLVC-Net~\shortcite{wen2021comprehensive}~\textcolor{gray}{(SIGIR'21)}& 17.64  & 54.39  & 79.47  & 50.50  \\
    ARTEMIS~\shortcite{delmas2022artemis}~\textcolor{gray}{(ICLR'22)} &  18.72  & 53.11  & 79.31  & 50.38 \\
    C-Former~\shortcite{xu2023multi}~\textcolor{gray}{(TMM'23)} & -     & 52.20  & 72.20  & - \\
    MGUR~\shortcite{chen2022composed}~\textcolor{gray}{(ICLR'24)}& 18.41  & 53.63  & 79.84  & 50.63  \\
    \hdashline
    \rowcolor[rgb]{ .949,  .949,  .949} \multicolumn{5}{c}{\textit{VLP Model-Based Methods}}\\
    FashionVLP~\shortcite{Goenka2022}~\textcolor{gray}{(CVPR'22)} & -     & 49.08  & 77.32  & - \\
    Prog. Lrn.~\shortcite{zhao2022progressive}~\textcolor{gray}{(SIGIR'22)}&  22.88  & 58.83  & 84.16  & 55.29  \\
    TG-CIR~\cite{tgcir}~\textcolor{gray}{(ACM MM'23)} & 25.89 &63.20 &85.07 &58.05 \\
    IUDC~\cite{iudc}~\textcolor{gray}{(TOIS'24)} & 21.17&  56.82&  82.25 &53.41\\
    LIMN~\cite{limn}~\textcolor{gray}{(TPAMI'24)} &- & 68.20 & 87.45 & -\\
    LIMN+~\cite{limn}~\textcolor{gray}{(TPAMI'24)} &- & 68.37 & 88.07 & -\\
    DQU-CIR~\shortcite{dqu}~\textcolor{gray}{(SIGIR'24)}& \underline{31.47} & \underline{69.19} & \underline{88.52} & \underline{63.06} \\
    ENCODER~\cite{encoder}~\textcolor{gray}{(AAAI'25)}& 26.97 &65.59& 86.48& 59.68\\
    \rowcolor[rgb]{ .851,  .851,  .851}
    \multicolumn{1}{l|}{\textbf{OFFSET(Ours)}} & \textbf{\large31.52} &	\textbf{\large69.96} &	\textbf{\large89.21} &	\textbf{\large63.56}  \\
\hline\hline
    \end{tabular}
}
      \label{tab:shoes}%
\end{table}%

\begin{table*}[ht!]
  \centering
  \vspace{-10pt}
      \caption{Performance comparison on CIRR with respect to R@$k$($\%$) and R$_{subset}$@$k$($\%$). The overall best results are in bold, while the best results over baselines are underlined.}
        \vspace{-10pt}
        \tabcolsep=6pt
      \resizebox{0.84\linewidth}{!}{
    \begin{tabular}{l|cccc|ccc|c}
\hline\hline
    \multicolumn{1}{c|}{\multirow{2}{*}{Method}} & \multicolumn{4}{c|}{R@$k$}      & \multicolumn{3}{c|}{R$_{subset}$@$k$} & \multirow{2}{*}{(R@5+R$_{subset}$@$1$)/$2$} \\
\cline{2-8}         & k=1   & k=5   & k=10  & k=50  & k=1   & k=2   & k=3   &  \\
    \hline
    \rowcolor[rgb]{ .949,  .949,  .949} \multicolumn{9}{c}{\textit{Traditional Model-Based Methods}} \\
    TIRG~\shortcite{vo2019composing}~\textcolor{gray}{(CVPR'19)}& 14.61  & 48.37  & 64.08  & 90.03  & 22.67  & 44.97  & 65.14  & 35.52  \\
    CIRPLANT~\shortcite{liu2021image}~\textcolor{gray}{(ICCV'21)}& 19.55  & 52.55  & 68.39  & 92.38  & 39.20  & 63.03  & 79.49  & 45.88  \\
    ARTEMIS~\shortcite{delmas2022artemis}~\textcolor{gray}{(ICLR'22)}& 16.96  & 46.10  & 61.31  & 87.73  & 39.99  & 62.20  & 75.67  & 43.05  \\
    C-Former~\shortcite{xu2023multi}~\textcolor{gray}{(TMM'23)}& 25.76  & 61.76  & 75.90  & 95.13  & 51.86  & 76.26  & 89.25  & 56.81  \\
    \hdashline
\rowcolor[rgb]{ .949,  .949,  .949} \multicolumn{9}{c}{\textit{VLP Model-Based Methods}}\\
    LF-CLIP~\shortcite{Baldrati_2022_CVPR}~\textcolor{gray}{(CVPR'22)}& 33.59  & 65.35  & 77.35  & 95.21  & 62.39  & 81.81  & 92.02  & 63.87  \\
    CLIP4CIR~\shortcite{baldrati2022conditioned}~\textcolor{gray}{(CVPRW'22)}& 38.53  & 69.98  & 81.86  & 95.93  & 68.19  & 85.64  & 94.17  & 69.09  \\
    TG-CIR~\cite{tgcir}~\textcolor{gray}{(ACM MM'23)} & 45.25  & 78.29  & 87.16  & 97.30  & 72.84  & 89.25  & 95.13  & 75.57  \\
    FashionERN~\cite{fashionern}~\textcolor{gray}{(AAAI'24)}  & -     & 74.77  & -     &  &  74.93  & -     & -     & 74.85  \\
    LIMN~\cite{limn}~\textcolor{gray}{(TPAMI'24)} &43.64  & 75.37  & 85.42  & 97.04  & 69.01  & 86.22  & 94.19  & 72.19 \\
    LIMN+~\cite{limn}~\textcolor{gray}{(TPAMI'24)} &43.33 &	75.41 &	85.81& 	97.21 &	69.28 &	86.43 &	94.26 	&72.35 \\
    SADN~\cite{sadn}~\textcolor{gray}{(ACM MM'24)} &44.27& 78.10& 87.71 &97.89& 72.71 &89.33& 95.38& 75.41\\
    DQU-CIR~\shortcite{dqu}~\textcolor{gray}{(SIGIR'24)}& 46.22 &	78.17 &	87.64 &	97.81 	&70.92 &	87.69 &	94.68 &	74.55  \\
    CaLa~\cite{cala}~\textcolor{gray}{(SIGIR'24)} &49.11 &	81.21 &	89.59 &	98.00 &	76.27 	&91.04 &	96.46 &	78.74  \\
    CoVR-2~\cite{covr-2}~\textcolor{gray}{(TPAMI'24)} &50.43 &81.08& 88.89& \underline{98.05}& 76.75& 90.34& 95.78 & 79.28\\
    Candidate~\cite{candidate}~\textcolor{gray}{(TMLR'24)}& 50.55 &	81.75 &	\underline{89.78} &	97.18 &	80.04 &	91.90 &	96.58 &	80.90 \\
    SPRC~\cite{sprc}~\textcolor{gray}{(ICLR'24)}& \underline{51.96} &	\underline{82.12} &	89.74 &	97.69 &	\underline{80.65} &	\underline{92.31} &	\underline{96.60} 	& \underline{81.39}  \\
    ENCODER~\cite{encoder}~\textcolor{gray}{(AAAI'25)}& 46.10& 77.98& 87.16 &97.64& 76.92& 90.41& 95.95 &77.45\\
    
    \rowcolor[rgb]{ .851,  .851,  .851}
    \multicolumn{1}{l|}{\textbf{OFFSET(Ours)}}  & \textbf{\large52.19} &	\textbf{\large82.60} &	\textbf{\large90.07} &	\textbf{\large98.07} &	\textbf{\large81.37} &	\textbf{\large93.08} 	&\textbf{\large97.54} & \textbf{\large81.99}  \\
\hline \hline
    \end{tabular}
}
\vspace{-10pt}
  \label{tab:cirr}%
\end{table*}%

\begin{table}[htbp]
  \centering
  \caption{Ablation Studies of OFFSET with various settings on FashionIQ, Shoes, and CIRR. $\Delta$ represents the performance degradation of the compared derivatives and is marked with \textcolor{defgreen4}{\textit{the green background}}. \textcolor{defyellowtext}{\textit{The yellow background}} denotes the baseline performance utilized for per column.}
  \vspace{-10pt}
    \resizebox{0.92\linewidth}{!}{
    \begin{tabular}{cl|cc|cc|cc}
    \hline
    \hline
    \multicolumn{1}{c|}{\multirow{2}{*}{D\#}} & \multicolumn{1}{c|}{\multirow{2}{*}{Derivatives}} & \multicolumn{2}{c|}{FashionIQ} & \multicolumn{2}{c|}{Shoes} & \multicolumn{2}{c}{CIRR} \\
\cline{3-8}    \multicolumn{1}{c|}{} &       & Avg.   & $\Delta$ & Avg.   & $\Delta$ & Avg.   & $\Delta$\\
    \hline
    \rowcolor[rgb]{ .949,  .949,  .949} \multicolumn{8}{c}{\textit{\textbf{G1:} Ablation on Dual Focus Mapping}}\\
    \multicolumn{1}{c|}{(1)} & w/o FM & 67.57  & \cellcolor{defgreen4}-4.82  & 59.28  & \cellcolor{defgreen4}-4.28  & 78.05  & \cellcolor{defgreen4}-3.94  \\
    \multicolumn{1}{c|}{(2)} & w/o VFM & 70.55  & \cellcolor{defgreen2}-1.84  & 62.15  & \cellcolor{defgreen2}-1.41  & 81.46  &  \cellcolor{defgreen1}-0.53  \\
    \multicolumn{1}{c|}{(3)} & w/o TFM & 70.00  & \cellcolor{defgreen3}-2.39  & 62.19  &  \cellcolor{defgreen3}-1.37  & 81.43  & \cellcolor{defgreen2}-0.56 \\
    \multicolumn{1}{c|}{(4)} & w/o MG-FP & 71.28  &  \cellcolor{defgreen1}-1.11  & 61.94  & \cellcolor{defgreen3}-1.62  & 78.54  & \cellcolor{defgreen3}-3.45 \\
    \hdashline
\rowcolor[rgb]{ .949,  .949,  .949} \multicolumn{8}{c}{\textit{\textbf{G2:} Ablation on Textually Guided Focus Revision}}\\
    \multicolumn{1}{c|}{(5)} & w/o Target\_VFM & 70.98  & \cellcolor{defgreen1}-1.41  & 61.70  & \cellcolor{defgreen2}-1.86  & 79.65  & \cellcolor{defgreen1}-2.34  \\
    \multicolumn{1}{c|}{(6)} & w/o Target\_MGFP & 70.48  & \cellcolor{defgreen2}-1.91  & 60.73  & \cellcolor{defgreen3}-2.83  & 78.59  & \cellcolor{defgreen2}-3.40  \\
    \multicolumn{1}{c|}{(7)} & w/o Revision & 69.05  & \cellcolor{defgreen3}-3.34  & 62.11  & \cellcolor{defgreen1}-1.45  & 78.49  & \cellcolor{defgreen3}-3.50  \\
    \hdashline
   \rowcolor[rgb]{ .949,  .949,  .949} \multicolumn{8}{c}{\textit{\textbf{G3:} Ablation on Optimization Functions}}\\
    \multicolumn{1}{c|}{(8)} & w/o BBC & 65.48  & \cellcolor{defgreen4}-6.91  & 30.14  & \cellcolor{defgreen4}-33.42  & 45.55  & \cellcolor{defgreen4}-36.44  \\
    \multicolumn{1}{c|}{(9)} & w/o FR & 71.08  & \cellcolor{defgreen2}-1.31  & 61.27  & \cellcolor{defgreen2}-2.29  & 78.10  & \cellcolor{defgreen2}-3.89  \\
    \rowcolor[rgb]{ .851,  .851,  .851}
    \multicolumn{2}{c|}{\textbf{OFFSET(Ours)}} & \textbf{72.39}  & \cellcolor{defyellow}0.00  & \textbf{63.56}  & \cellcolor{defyellow}0.00  & \textbf{81.99}  & \cellcolor{defyellow}0.00  \\
    \hline
    \hline
    \end{tabular}%
    }
    \vspace{-13pt}
  \label{tab:ablation}%
\end{table}%

\subsection{Performance Comparison}
 We compare OFFSET with several CIR methods, which can be classified into two categories according to the utilized backbone: traditional model-based baselines (\textit{e.g.}, TIRG~(\cite{vo2019composing}), MGUR~(\cite{chen2022composed})) and CLIP-based baselines (\textit{e.g.} DQU-CIR~(\cite{dqu}), ENCODER~(\cite{encoder})).
Our analysis of the comparative data in Table~\ref{tab:fiq_shoes}, Table~\ref{tab:shoes}, and Table~\ref{tab:cirr} yields the following significant observations: 
\textbf{1)} OFFSET consistently outperforms all baseline models on FashionIQ, Shoes, and CIRR datasets. Specifically, OFFSET achieves $1.00$\% relative improvements over the best baseline for R@$10$ on FashionIQ-Avg, $0.89$\% for R$_{subset}$@$1$ on CIRR, and $0.79$\% for average on Shoes, respectively, demonstrating its effectiveness and generalization ability in both fashion-specific and open-domain CIR tasks. 
\textbf{2)} The VLP-based models (bottom of tables) typically outperform those based on traditional feature extraction models (top of tables), which confirms the effectiveness of VLP-based models applied to the CIR task and provides a solid foundation for visual-textual semantic alignment.
\textbf{3)} DQU-CIR outperforms all other baselines on all metrics in the fashion-domain dataset, but its performance on the open-domain dataset CIRR remains inferior. This may be due to the complexity of the open-domain dataset, which hinders the model's OCR capability to accurately recognize the semantics of multimodal queries' keywords, thus limiting its performance. In contrast, OFFSET not only outperforms the previous models in fashion-domain datasets but also achieves state-of-the-art results on the open-domain dataset CIRR, which indicates that OFFSET has stable CIR performance and is not limited to domain-specific data.

\subsection{Ablation Studies}
To illuminate the pivotal role of each module and optimization function in our proposed model OFFSET, we conducted in-depth comparisons among OFFSET and its derivatives, which can be classified into three groups as follows.

\textit{\textbf{$\bullet$ G1:Ablation on Dual Focus Mapping.}} This group aims to validate the effectiveness of the modules in \textit{Dual Focus Mapping}. Specifically, the compared derivatives in this group are as follows.
\textbf{D\#(1) w/o FM, D\#(2) w/o VFM} and \textbf{D\#(3) w/o TFM}: To validate the effect of Dual Focus Mapping in OFFSET, we separately remove VFM and TFM.
\textbf{D\#(4) w/o MG-FP}: To explore the impact of the Multi-Grained Focus Projection module when aligning global and local focus semantics, we replace MG-FP with the simple average pooling to obtain the projected features.

From the results of \textbf{\textit{G1}} in Table~\ref{tab:ablation}, we can obtain the following observations. 
\textbf{1)} \textbf{D\#(1) w/o FM} performs the worst among the variants in this group. This demonstrates the necessity of simultaneously enabling the visual and
textual features to focus on the portions that are closely related to the multimodal queries.
\textbf{2)} both \textbf{D\#(2)} and \textbf{D\#(3)} are inferior to OFFSET, which indicates that independently performing focus mapping for visual and textual features can also effectively focus their portions close to the semantics closely related to the multimodal queries.
\textbf{3)} The degradation in performance of \textbf{D\#(4)} reveals the importance of \textit{Multi-Grained Focus Projection} in precisely aligning the local and global focus.

\textit{\textbf{$\bullet$ G2:Ablation on Textually Guided Focus Revision.}} 
This group is designed to demonstrate the validity of the utilized modules in \textit{Textually Guided Focus Revision}. Concretely, the compared derivatives in this group are as follows.
\textbf{D\#(5) w/o Target\_VFM} and \textbf{D\#(6) w/o Target\_MG-FP}: To validate the necessity of applying the VFM and MG-FP modules to the target image during \textit{Textually Guided Focus Revision}, we remove VFM and MG-FP on the target image in these two variants, respectively.
\textbf{D\#(7) w/o Revision}: To assess the efficacy of textually guided focus revision, we replace the revision process with a simple addition operation between the focus features.

From the experimental results of \textbf{\textit{G2}} in Table~\ref{tab:ablation}, the following findings can be obtained.
\textbf{1)} both \textbf{D\#(5)} and \textbf{D\#(6)} demonstrate a performance decline compared to the complete OFFSET. This indicates the necessity of performing \textit{Visual Focus Mapping} and \textit{Multi-Grained Focus Projection} on the target image, which makes the target semantics of focused portions more close to the semantics of multimodal queries.
\textbf{2)} \textbf{D\#(7)} exhibits a significant gap compared to OFFSET, demonstrating that the focus revision process is indeed effective for revising the focus of the multimodal query. 

\textit{\textbf{$\bullet$ G3:Ablation on Optimization Functions}} This group is devised to explore the effect of the optimization functions of OFFSET, whose derivatives are listed as follows.
\textbf{D\#(8) w/o BBC}: To check the effect of the batch-based classification loss (BBC, Eqn.$($\ref{bbc}$)$), we remove $\mathcal{L}_{rank}$ in Eqn.$($\ref{optimization}$)$.
\textbf{D\#(9) w/o FR}: To validate the impact of focus regularization (FR, Eqn.$($\ref{L_fr}$)$), we ablate $\mathcal{L}_{fr}$ in Eqn.$($\ref{optimization}$)$.

We can observe from the results of \textit{\textbf{G3}} in Table~\ref{tab:ablation} that:
\textbf{1)} \textbf{D\#(8)} performs worse than OFFSET, which proves the effectiveness of the batch-based classification loss in guiding the model to learn better multimodal focus features. 
\textbf{2)} The performance drop of \textbf{D\#(9)} indicates that focus regularization plays a vital role in maintaining focus consistency between the multimodal query and target.

\subsection{Further Analysis}

In this section, to further demonstrate the effectiveness of OFFSET, we test the parameter sensitivity of OFFSET to the focus channel number $P$ and the inference efficiency comparison with the representative CIR model, DQU-CIR~\cite{dqu}. Moreover, we exhibit the qualitative results of OFFSET to visually illustrate its retrieval performance.
We describe the experimental results in detail as follows.

\subsubsection{Sensitivity to Focus Channel Number $P$}

\begin{figure}[ht]
\centering
\vspace{-8pt}
\includegraphics[width=\linewidth]{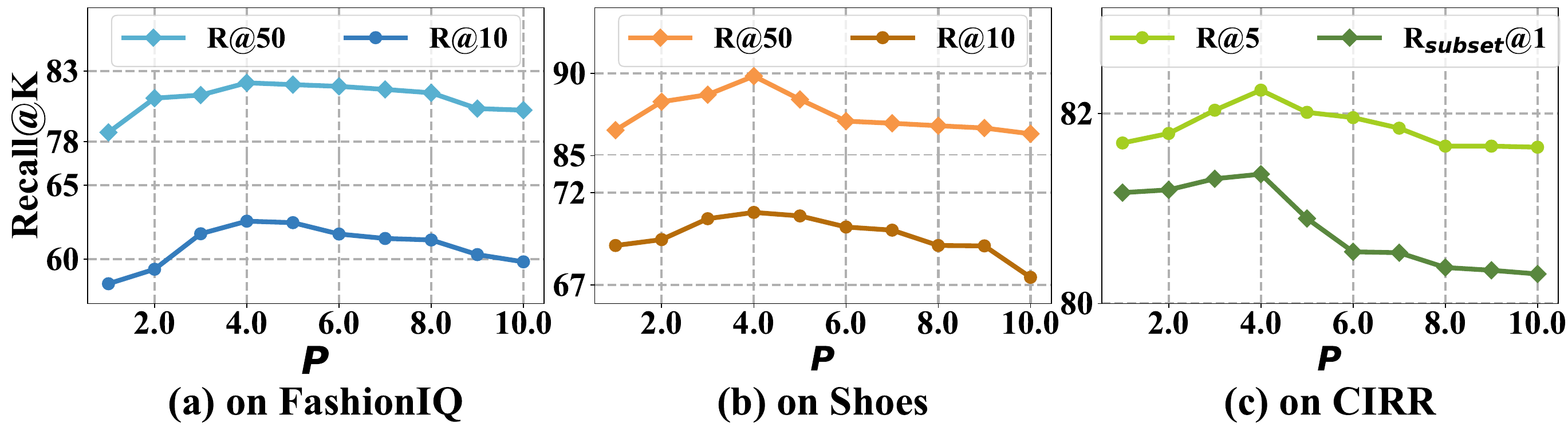}
\vspace{-22pt}
   \caption{Sensitivity to Focus Channel Number $P$ and the hyper-parameter $\mu$ on (a) FashionIQ, (b) Shoes, and (c) CIRR.}
\label{fig:chanel_num}
\vspace{-10pt}
\end{figure}

To investigate the sensitivity of OFFSET to the focus channel number $P$, 
we present performance comparison with various $P$ on FashionIQ, Shoes, and CIRR datasets in Figure~\ref{fig:chanel_num} (a)-(c), respectively.
From the figure, we observe that the performance of OFFSET generally improves as the focus channel number $P$ increases, then drops for larger values. This is reasonable since a certain number of focus channels is necessary to effectively capture diverse aspects of the image features. However, too many channels may cause irrelevant information to be focused on, hence disadvantaging the retrieval performance.

\subsubsection{Efficiency Analysis}
In Table~\ref{tab:inference}, we present a comparison of the inference efficiency between the proposed OFFSET and the representative CIR model DQU-CIR~\cite{dqu}. Specifically, we list the inference time per sample, the corresponding retrieval performance on FashionIQ and CIRR, and additional auxiliary models used by the two models (\textit{i.e.,} the caption model (BLIP-2~\cite{li2023blip} for both DQU-CIR and OFFSET), the segmentation model (CLIPSeg~\cite{luddecke2022image} for OFFSET), and the LLM (Gemini-pro-v1 for DQU-CIR)). All experiments are conducted on a single A40 GPU. From the results in the table, we observe that the inference time of our proposed OFFSET decreases by $40.27$\% compared to DQU-CIR, while the retrieval performance on all datasets outperforms that of DQU-CIR (as illustrated in Table~\ref{tab:fiq_shoes}, Table~\ref{tab:shoes}, and Table~\ref{tab:cirr}), especially with an improvement of $9.98$\% on CIRR-Avg. This indicates that OFFSET achieves optimal retrieval performance without incurring excessive additional overhead, which aligns with the retrieval intent of the CIR task.

\begin{table}[htbp]
  \centering
  \vspace{-8pt}
  \caption{Comparison of inference efficiency. The better results are in bold. Cap. represents the utilized caption model and Seg. denotes the utilized segmentation model. F-R10-Avg. is the average of R@$10$ on FashionIQ and C-Avg. represents the ((R@$5$+R$_{subset}$@$1$)/$2$) on CIRR.}
  \vspace{-10pt}
  \resizebox{\linewidth}{!}{
    \begin{tabular}{c|c|c|c|c|c|c}
    \Xhline{1pt}
    Methods & Cap. & Seg. & LLM   & Test$\downarrow$  & F-R10-Avg.$\uparrow$ & C-Avg.$\uparrow$ \\
    \hline
    DQU-CIR & BLIP-2 & \usym{1F5F6} & Gemini & 2.05s & 71.77& 74.55 \\
    \rowcolor[rgb]{ .851,  .851,  .851}
    OFFSET & BLIP-2 & CLIPSeg & \usym{1F5F6} & \textbf{1.04s} & \textbf{72.39} &\textbf{81.99} \\
    \Xhline{1pt}
    \end{tabular}%
    }
    \vspace{-8pt}
  \label{tab:inference}%
\end{table}%

\subsubsection{Qualitative Analysis}

\begin{figure}[h]
\centering
\vspace{-9pt}
\includegraphics[width=0.95\linewidth]{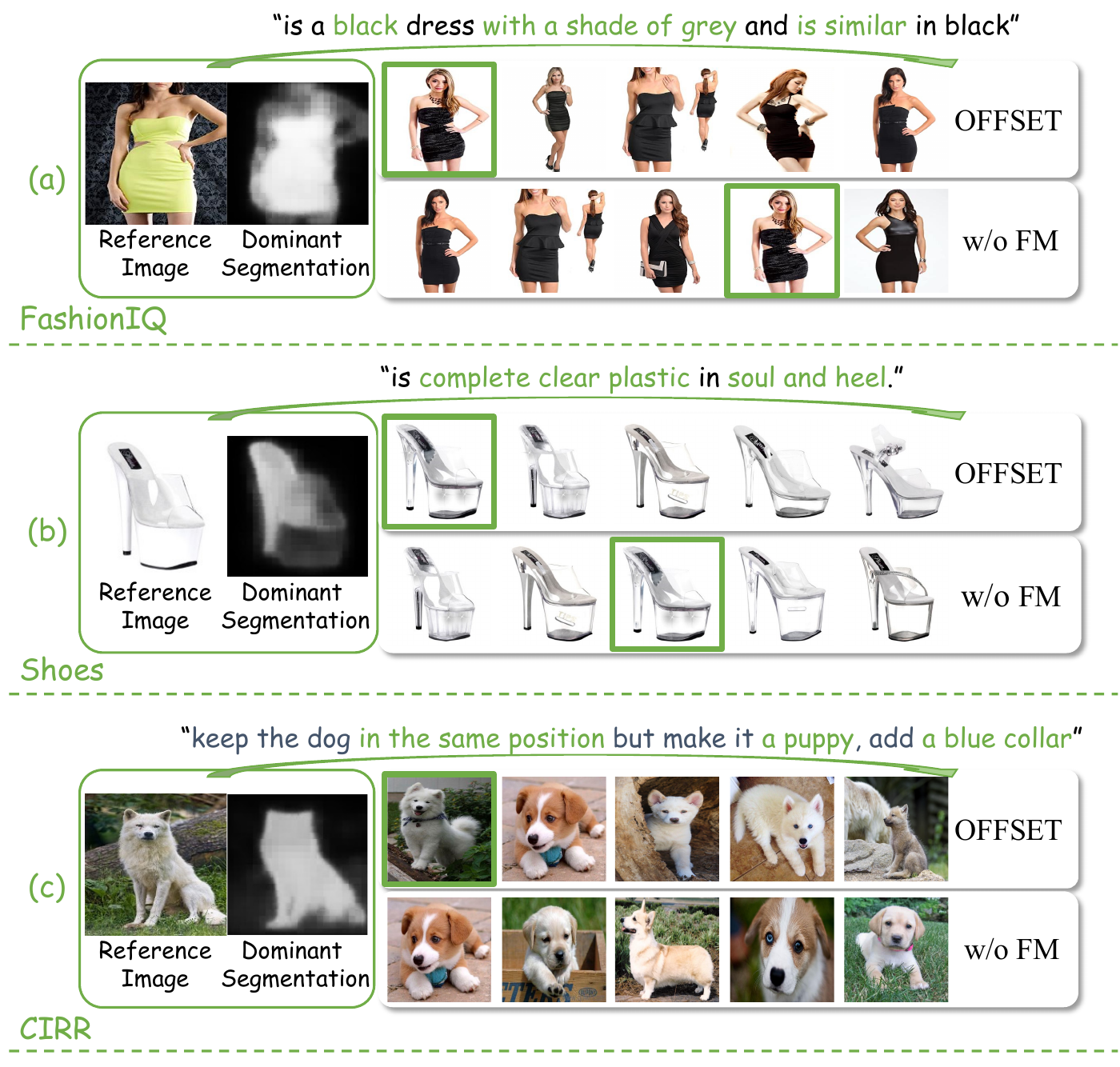}
\vspace{-14pt}
   \caption{Case study on (a) FashionIQ and (b) CIRR.}
\label{fig:attention}
\vspace{-15pt}
\end{figure}

Figure~\ref{fig:attention} displays the top 5 retrieved images of two CIR examples obtained by OFFSET and its derivative w/o FM on the fashion-domain FashionIQ, Shoes and the open-domain CIRR datasets. The green boxes indicate the target images. As shown in Figure~\ref{fig:attention}, OFFSET successfully ranks the target image in the first place on the three datasets, while w/o FM fails and even ranks it out of the top 5 on the CIRR dataset. 
Meanwhile, we observe that OFFSET can maintain the unmodified portion (\textit{e.g.} the vacancy around the dress waist in (a)) and capture the nuanced requirements specified in the text (\textit{e.g.}, ``the same position'' and ``a blue collar'' in (c)) more accurately than w/o FM. 
In addition, OFFSET are capable of comprehending the detailed descriptions in the modification text. As illustrated in Figure~\ref{fig:attention}(b), OFFSET successfully recognizes the image in which both the heel and sole are transparent, while w/o FM focuses on the sole only.
These results suggest that OFFSET better interprets the relationship between the reference image and the modification text, leading to more precise retrievals that match the user's intent.

\section{Conclusion}
This work found that the CIR community has two phenomena, which is seriously neglected:
1) the inhomogeneity in visual data leads to the degradation of query features. 2) the text-priority in multimodal queries leads to visual focus bias. In light of these findings, we proposed OFFSET, which encompasses three advantages. Firstly, to address the inhomogeneity, we developed a focus mapping-based feature extractor, which identifies the dominant region and guides the visual and textual feature extraction, thereby reducing noise interference. Secondly, for the phenomenon of text-priority, we proposed textually guided focus revision, which enables the adaptive focus revision on the reference image according to the modification semantics, thus enhancing the perception of the modification focus on the composed features. Ultimately, extensive experiments on four benchmark datasets substantiated the efficacy of our proposed OFFSET. In the future, we intend to extend our approach to address other downstream tasks, such as information detection and prediction~\cite{wu2025k2vae,qiu2024tfb,liu2025rethinking,qiu2025duet}.

\bibliographystyle{ACM-Reference-Format}
\bibliography{reference}

\end{document}